\documentclass[journal, usletter]{ieeeconf}  

\IEEEoverridecommandlockouts                              

\usepackage{amsmath} 

\usepackage{graphicx}
\usepackage{comment}
\usepackage{float}
\usepackage{hyperref}
\usepackage{booktabs}

\title{\LARGE \bf
EgoNav: Egocentric Scene-aware Human Trajectory Prediction
}

\author{Weizhuo Wang$^{1}$, C. Karen Liu$^{1}$, and Monroe Kennedy III$^{1}$
\thanks{$^{1}$Stanford University, CA 94305, USA}
\thanks{\tt\small\{weizhuo2, ckliu38, monroek\}@stanford.edu.}
}

\begin{document}


\maketitle
\thispagestyle{empty}
\pagestyle{empty}

\begin{figure*}[tb]
  \centering
  \includegraphics[width=\linewidth]{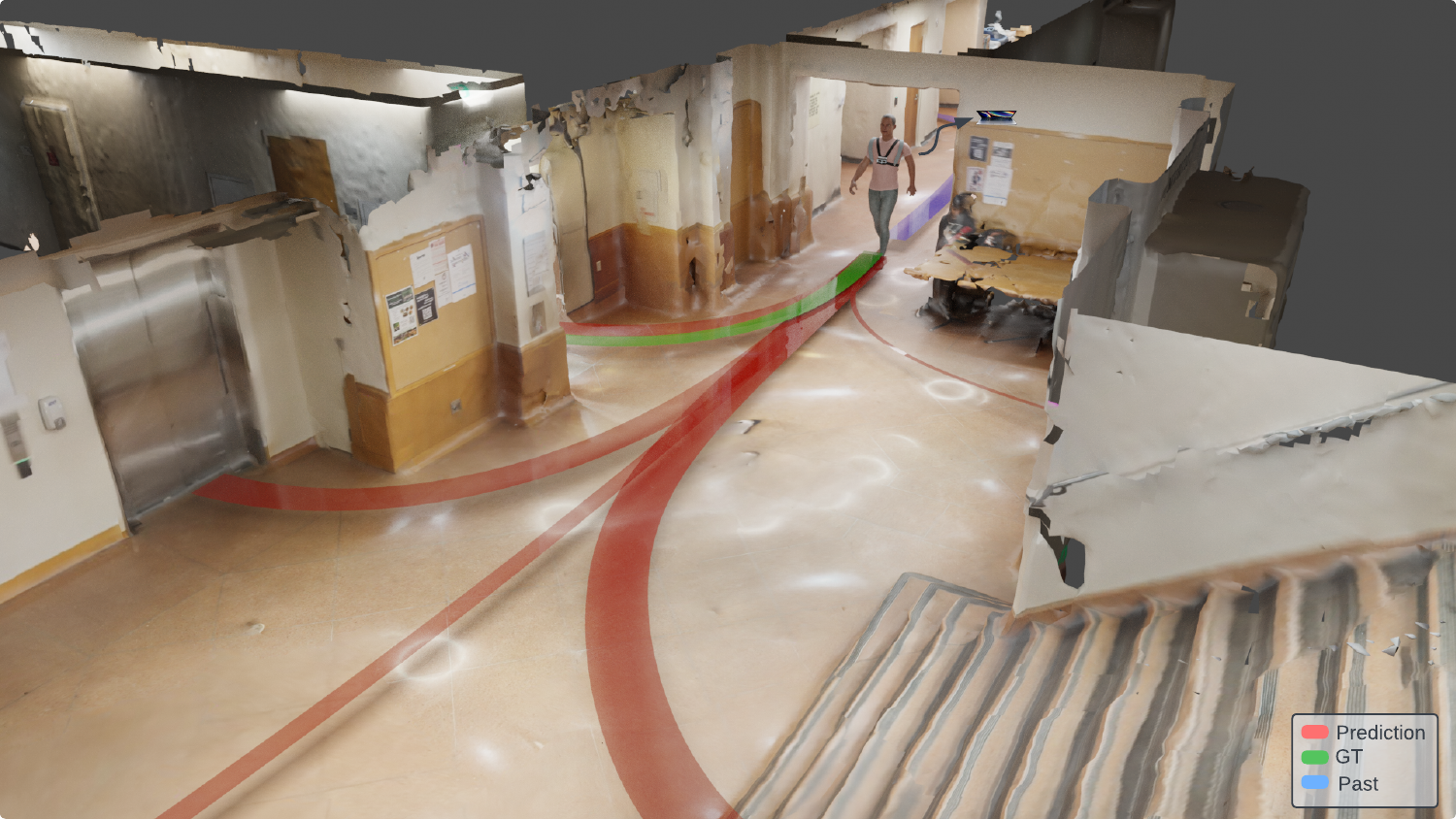}
  \caption{\textbf{Visual Representation of the Problem:} Given the past in blue, and learning from single-modal ground truth in green, what are the likely future paths and their likelihood? Different red ribbon illustrates different possible modes in the scene and the size of the ribbon denotes the likelihood. }
  \label{fig:outline}
\end{figure*}

\begin{abstract}
    Wearable collaborative robots stand to assist human wearers who need fall prevention assistance or wear exoskeletons. Such a robot needs to be able to constantly adapt to the surrounding scene based on egocentric vision, and predict the ego motion of the wearer. In this work, we leveraged body-mounted cameras and sensors to anticipate the trajectory of human wearers through complex surroundings. To facilitate research in ego-motion prediction, we have collected a comprehensive walking scene navigation dataset centered on the user's perspective. We then present a method to predict human motion conditioning on the surrounding static scene. Our method leverages a diffusion model to produce a distribution of potential future trajectories, taking into account the user's observation of the environment. To that end, we introduce a compact representation to encode the user's visual memory of the surroundings, as well as an efficient sample-generating technique to speed up real-time inference of a diffusion model. We ablate our model and compare it to baselines, and results show that our model outperforms existing methods on key metrics of collision avoidance and trajectory mode coverage. 
\end{abstract}

\section{Introduction}
\label{sec:intro}

The integration of autonomous systems into human-centric environments, particularly through collaborative robotics, necessitates acute awareness and prediction of human motion. These systems, whether as large as autonomous vehicles or wearable like exoskeletons, require precise motion prediction to ensure safety and enhance collaboration. Their successful operation relies on a foundation of continuously inferring the intention of human cooperators, through the same egocentric perspective.

Many studies \cite{chen_scept_2022, sun_large_2024, mozaffari_trajectory_2023, zhang_predictive_2023, vedaldi_pip_2020, karkus_diffstack_2022} suggest that there is a close relationship between trajectory predictor and motion planner. Traditionally, a downstream motion planning module utilizes trajectory predictions as input to generate plans within given constraints. Recently, imitation learning approaches \cite{zare_survey_2023} have emerged as an alternative for motion planning. These approaches learn a probability distribution of possible actions given observations, which fundamentally relies on predicting future trajectories from the agent's perspective. This egocentric prediction is crucial for imitation learning, as it allows the agent to anticipate and replicate human-like behavior in various scenarios. Notably, both paradigms - separate planners and imitation learning methods - can significantly benefit from a robust egocentric trajectory prediction module. This paper proposes a novel egocentric human trajectory prediction model that not only advances the state of the art in prediction but also serves as a valuable component for both traditional planning systems and imitation-based approaches. In the context of embodied agents, our model represents a crucial step towards creating more natural and adaptive motion planning and control systems, regardless of the specific planning paradigm employed.

Predicting motion involves deciphering the complex relationship between past movements, scene semantics, and potential future actions. This task is complicated by the need to account for multi-modal outcomes, for instance, a person could have many equally viable choices at a divergence. Factors influencing these decisions range from high-level objectives to immediate environmental constraints like obstacles and walkable paths. Previously, work like Social Force  \cite{helbing_social_1995} attempted to solve this task through a series of non-linearly coupled equations and simplified attracting repelling dynamics. Recently learning-based solutions such as \cite{wang_trajectory_2023, singh_krishnacam_2016} have leveraged the power of data to implicitly model the more complicated preferences. To further improve upon the limitations, especially in scene understanding, we propose to use semantic segmentation to add dense channel labels to the scene. So that the model can relate different behaviors and preferences back to those labels.

Many previous work \cite{helbing_social_1995, cui_multimodal_2019, varadarajan_multipath_2022} assume the environment to be fully observed, notably in a bird's eye view (BEV). However, a fully known environment is rarely obtainable in the real world, limiting the applicability of such algorithms. The egocentric perspective has much higher availability but also brings a unique set of challenges. The first is occlusion, especially when the scene is highly three-dimensional and cluttered. Second, a stereo camera creates a huge amount of raw data, unlike bird's eye view, it also has a limited field of view. These challenges require careful consideration that existing literature has not done a good job of tackling. That said, semantic segmentation and generative models are critical to solving this inherently multi-modal, and highly environment conditioning task. 

Unlike autonomous vehicles, the walking environment is highly unstructured and has very few traffic rules to narrow down the solution space. Most work \cite{wiest_probabilistic_2012, gu_stochastic_2022, lv_learning_2024} fails to consider the environmental features and only predicts based on the past trajectory. The pinnacle of the challenge is to solve egocentric, unstructured scene navigation with interactions with other agents in the scene. However, for now, there still remains a lot of work to do without interaction.

In this paper, we propose a generative modeling approach along with the environment semantics to predict the distribution of future trajectories of a person, given the egocentric view, as illustrated in Fig \ref{fig:outline}. To address the under-constrained and multi-modal nature of the navigation problem, we employ a diffusion model to predict the future trajectories conditioned on the egocentric observation of the environment and the user's past walking trajectory. To account for numerous challenges brought by the egocentric view, we introduce a compact representation of the environment to capture a short history of visual observations from the egocentric perspective. This "visual memory" representation encodes both the appearance and semantics information of the world. To achieve real-time inference, we introduce a hybrid generation technique that balances between quality and speed of sample generation from the trained diffusion model. Using our hybrid generation method, we can rapidly sample multiple future trajectories in real-time, providing downstream applications a distribution of the possible paths a person might take, rather than predicting a single path.

We evaluate our predictive model using real-world navigation data. We collect a comprehensive walking scene navigation dataset centered on the user's perspective in diverse indoor and outdoor scenarios. We show that our model can accurately predict a distribution of future trajectories representing human preference taking scene context into account, outperforming the baseline methods on various evaluation metrics. We also conduct ablation studies to validate the design choices in visual memory representation and hybrid generation technique. The dataset and the trained model will be made publicly available. 



\section{Related Work}
\label{sec:relatedwork}
Path prediction of humans is an area of interest across the robotic and computer vision fields. In this section, we highlight key works in various domains such as autonomous vehicles, pedestrians walking directions, motion during sports activities, and humans in social settings leveraging social context. We also highlight recent advances in diffusion models as a method of generating various outputs.
\begin{figure*}[tb]
  \centering
  \includegraphics[width=\linewidth]{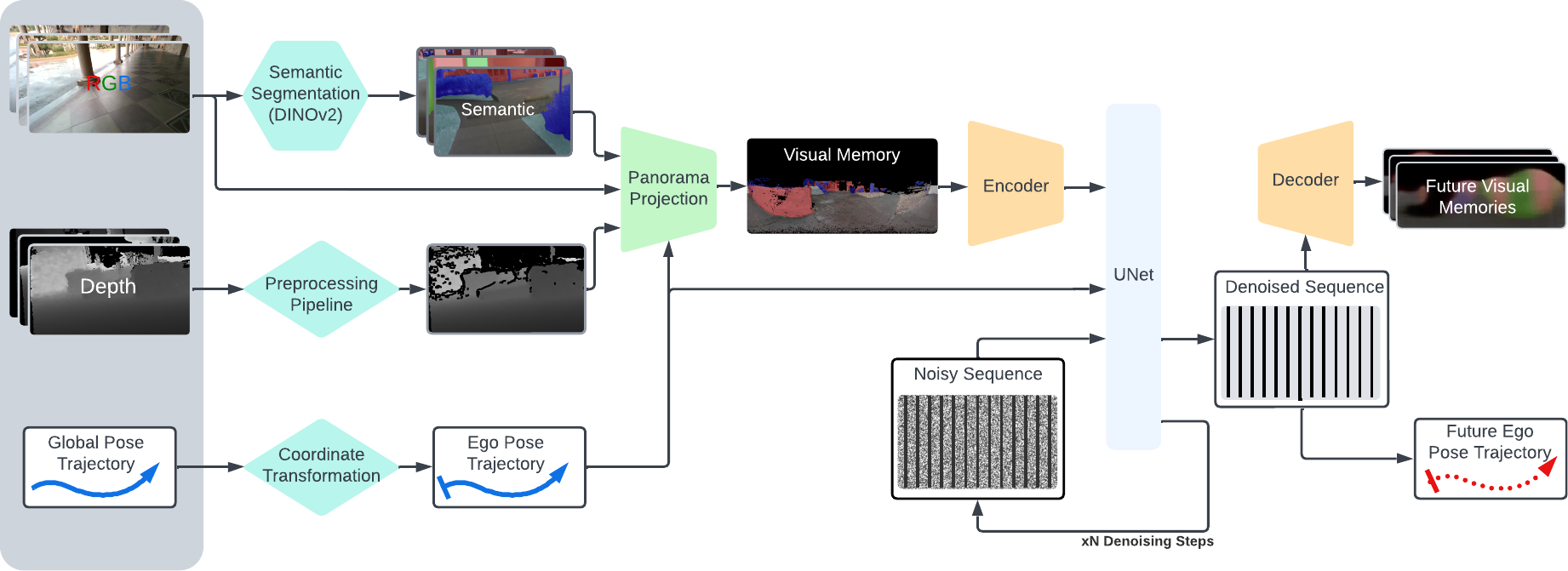}
  \caption{\textbf{Overview of the proposed method:} We maintain a 5-second buffer of logs that is most relevant to the prediction and organize them into a visual memory frame. All input and output of the prediction module are in the ego-centric frame.}
  \label{fig:sys_overview}
\end{figure*}

For autonomous vehicles (AV), predicting the motion of both ego-vehicle and other traffic agents is crucial for the AV to plan safe, informed trajectories \cite{rempe_trace_2023}. Albeit the problem setup can vary from our task, sometimes more emphasizing third-person view trajectory prediction, we can still learn from their attempts. Recent approaches focus on extracting motion prediction from demonstrations \cite{karkus_diffstack_2022}, and integrate directly with planner in a differentiable manner. Traditionally algorithms for trajectory prediction relied on recurrent models such as RNNs, LSTMs, or GRUs  \cite{tang_multiple_2019}. However, these models show limitations when confronted with complex, multi-modal sensor data, mainly due to the hidden state size and forgetting information in long sequences. Transformer-based models provide a powerful alternative offering benefits in multi-modal sensor streams and the ability to handle long-sequence inter-dependence's \cite{mercat_multi-head_2019, vaswani_attention_2023}. For AVs, with numerous guiding traffic rules, it is often convenient to reconstruct the environment from a birds-eye-view (BEV) or 2D occupancy grid to fuse sensory inputs \cite{strohbeck_multiple_2020, schreiber_long-term_2019, chai_multipath_2019}. While such representations are easy to interpret and predictions are limited to a small set of modes or choices, they introduce a "long tail" problem which is particularly troublesome when the structured environment assumption crumbles, presenting a need for research in representations that balance structure and flexibility in trajectory prediction.\cite{rempe_generating_2022}  

Trajectory prediction of players during sports is another field of interest \cite{bertasius_egocentric_2018}. The task in this field is more focused on game strategy and interactions, and less on the environmental features. Thus sporting scenarios usually take place where there are a few, if any, static objects/obstacles, and playing grounds are usually flat, leaving game strategy as the sole motion intent informer \cite{su_predicting_2017, bertasius_egocentric_2018}.

Trajectory prediction of humans in walking settings is the domain this work most interested in. As mentioned in Sec.\ref{sec:intro}, initial attempts usually ignore the environmental features (or assume them to be fully known), and operate on a highly abstracted level. For example Social Force \cite{helbing_social_1995} models all agents as a particle and moves based on various attractive and repelling forces. More recent examples are Social GAN \cite{gupta_social_2018}, and Social LSTM \cite{alahi_social_2016}, which assume an open space with only pedestrians. This series of work including some more recent ones \cite{wiest_probabilistic_2012, gu_stochastic_2022, lv_learning_2024} have taken the interaction modeling to its extreme, they operate by analyzing only past trajectories and predicts future paths where behaviors are purely conditioned on social norms and settings. So far, such models have achieved some promising results in terms of prediction quality running offline. However, the complete overlook of environmental interactions and overly ideal assumptions have hindered the real-world applicability of such a model. Prominent datasets facilitating these studies include UCY/ETH \cite{lerner_crowds_2007} and the Stanford Drone Dataset\cite{robicquet_learning_2016}, which predominantly offer a BEV of pedestrian movement, simplifying the trajectory prediction challenge to a two-dimensional problem. This top-down perspective allows for the tracking of pedestrians through bounding boxes, offering an effective means to study and model pedestrian trajectories in a simplified setting.

Realizing that those problems have been preventing the deployment in the real world, the latest work has focused more on the perception side. Works in autonomous vehicle community \cite{cui_multimodal_2019, varadarajan_multipath_2022, chen_scept_2022, sun_large_2024} have first adopted synthesized BEV as a structured fusion of environmental cues. However, the less structured walking scenarios demand a more flexible representation than BEV. So far, there are some early work \cite{singh_krishnacam_2016, park_egocentric_2016} using first-person views RGB(D) as input. More recent work has attempted more sophisticated fusing methods such as depth panorama \cite{wang_trajectory_2023}, and pose estimation \cite{qiu_egocentric_2022} for more informative representation. We believe there is still potential yet to be leveraged in this direction.


The diffusion model, inspired by the physical diffusion process, was originally proposed in 2015 \cite{sohl-dickstein_deep_2015} but gained significant traction with the introduction of Denoising Diffusion Probabilistic Models (DDPM) in 2020 \cite{ho_denoising_2020}. Unlike initial attempts that struggled to predict denoised inputs directly, DDPM introduced an approach that predicts the noise to be removed, marking a breakthrough in model performance. Subsequent developments, such as Latent Diffusion Models \cite{rombach_high-resolution_2022} for complex generation tasks, DDIM \cite{song_denoising_2022} for accelerated processing, and various guidance and conditioning techniques \cite{lugmayr_repaint_2022}, have expanded the application of diffusion models to image, video, and action generation. These models excel in generating detailed and coherent outputs across a range of generative tasks, proven by many successes in image, video, and even action-generation tasks. 

\section{Method}
\label{sec:method}

 \begin{figure*}[tb]
  \centering
  \includegraphics[width=\linewidth]{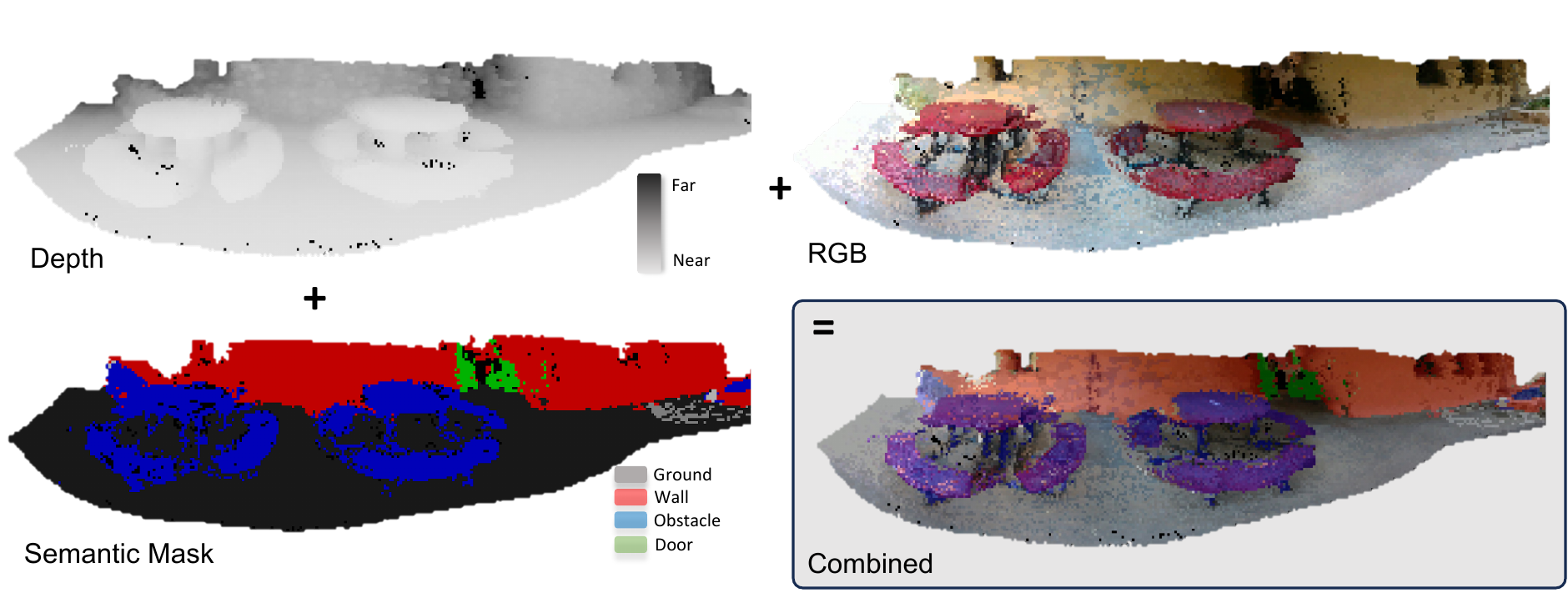}
  \caption{\textbf{Channels in Visual Memory:} The visual memory integrates frames from various time steps into a single panorama. It consists of a depth channel, color channel, and intensity-encoded 8-class semantic channel. 4 channels are shown in the figure.}
  \label{fig:panorama}
\end{figure*}

\subsection{Problem Formulation}
The goal of this work is to predict the possible paths of a person in a cluttered environment. A trajectory $\tau$ is a sequence of 6D poses (translation and orientation) of a person navigating in the 3D world. At each time step $t$, our model uses the past trajectory $\tau_{1:t}$ to predict likely future trajectories $\tau_{t:t+T}$. In addition, the prediction must be conditioned on the observation of surroundings.  The visual observation $S$ encodes the appearance, geometry, and semantics of the environment captured by wearable visual and depth sensors. Our goal is to model the probability distribution of the future trajectories: 

\begin{equation}
    \hat{\tau}_{t:t+T} \sim p_{\theta}(\cdot | \tau_{1:t},S)
    \label{eqn:prob_statement}
\end{equation}

We learn the model $p_{\theta}$ that best approximates the future trajectory, recognizing that the expected path may be sampled from a multi-modal distribution given the environment. An example of this would be a fork in the path, both paths may be viable, and multiple sampling from the model output distribution should recognize the distributed likelihood. 




\subsection{Method Overview}

Our method takes as input the past trajectory of the person recorded by an Intel Realsense T265 camera and a short history of RGBD images recorded by an outward-facing stereo camera worn by the person and predicts the future trajectories the person might take (Figure \ref{fig:sys_overview}. The color images are semantically labeled by DINOv2 \cite{oquab_dinov2_2024}, while the depth images go through a preprocessing pipeline that filters out erroneously filled edges. We transform the past trajectory from a global coordinate frame to an egocentric frame defined by the +Z as opposite to the gravity vector and forward-facing direction as +X. Taking the semantically labeled images, RGB images, and processed depth images, we project and align the images to create a single panorama in the egocentric coordinate frame, referred to as "visual memory". Conditioning on the visual memory and the past ego trajectory, a diffusion model is trained to predict the future trajectory along with encoded visual observations, as auxiliary outputs. The diffusion model starts from a random noise and performs a number of denoising steps to generate the clean prediction sequence. The clean sequence is a concatenation of encoded visual memory and future ego pose trajectory. Finally, we use the VAE decoder to recover the expected future panorama.

\begin{figure*}[tb]
  \centering
  \includegraphics[width=\linewidth]{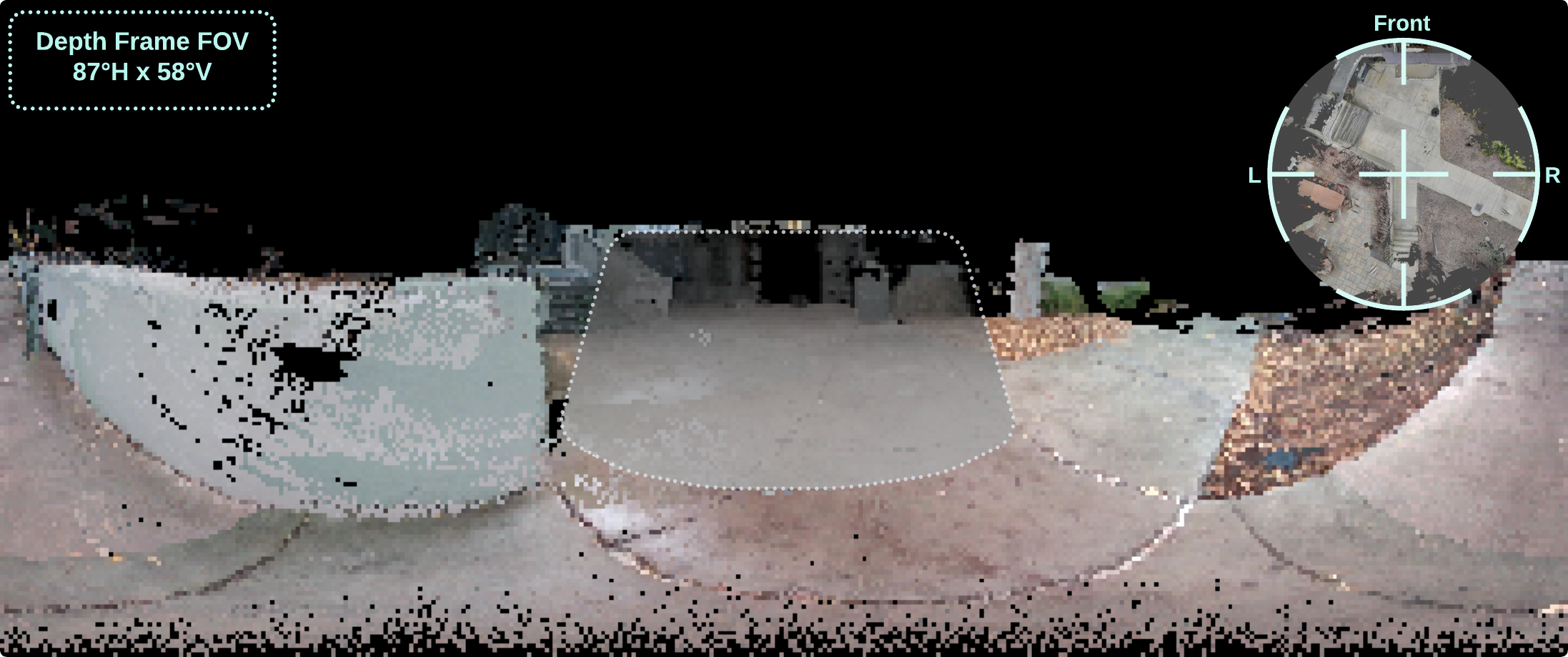}
  \caption{\textbf{Comparing depth frame with visual memory:} A raw depth frame from a stereo camera has only ~90 degrees of narrow FOV and often misses important scene information. In the figure, the depth frame only sees the open space in front and does not capture the stairs, the right turn path, or the wall directly to the left. The black regions are the undiscovered areas when stitching frames from different time steps.}
  \label{fig:semantic}
\end{figure*}

\subsection{Construction of Visual Memory}

Visual memory is defined as an ego-perspective, panoramic view representation of the surroundings. The visual memory is constructed from aligned color, depth, and semantic images (Shown in Fig \ref{fig:panorama}). Given the camera's intrinsic and extrinsic parameters, images in different frames can be projected to a single point cloud in the global frame. A distance-based filter is applied to remove points too far away from the current pose. The points are then projected back to the current ego frame to form a coherent representation of all the scene information gathered.

It is important to note that the visual memory representation holds a lot more relevant information than just a single image. As shown in Fig \ref{fig:semantic}, A single image from a stereo camera only has a narrow FOV pointing directly in front. It fails to capture the objects and paths in the scene that are highly relevant to the prediction, requiring many individual frames to be sent to the prediction module, relying on the model's capability to extract useful information. Meanwhile, the visual memory stitches the past frames together and integrates multi-modal inputs all into one single image, greatly improving the model and storage efficiency.

\subsection{Diffusion Pipeline for Trajectory Prediction}

The prediction part of the method heavily relies on a UNet diffusion model with multi-head self-attention layers between blocks. The self-attention layers help to relate the visual memory encoding to different parts of the trajectory, thus facilitating a deeper understanding of how humans interact with environmental features. 

The prediction is conditioned on a single 64-dimensional encoded visual memory, and full 100 steps of 6d pose trajectory at 20 Hz. We compress the Visual Memory (VM) from 1x180x360x5 (R-G-B-D-Segmentation) to a 1x64 dimensional embedding using a pre-trained VAE (Encoder). State trajectories are only normalized (100x9) since they are already small. The 9 channels represent x-y-z positions and ortho6d\cite{zhou_continuity_2020} orientation.

We utilize the full 100 steps as we believe in some cases the ego-centric trajectory prediction task can violate pure Markovian assumption. For example, a different past trajectory could likely affect the visibility of the scene in visual memory, causing different exploration preferences in trajectory selection. To verify the effectiveness, we compared it to a pure Markovian assumption in the ablation section.

The UNet takes VM embeddings and normalized states as conditions (Fig. \ref{fig:model_arch}). The conditions are integrated by classifier-free guidance, removing the input conditions 10\% of the training time. We apply both conditions to all down, up, and self-attention blocks, transforming the shape through a linear layer and adding to the output activation.

The output of the diffusion model is a denoised sequence that contains 100 steps of both normalized ego-centric trajectory and visual memory encoding, resulting in a (100, 9+64) sized prediction for all 100 steps. The normalized trajectory prediction is the first 9, and the embedded VM prediction is the rest. We use the same pre-trained VAE (Decoder) to decode the full 100x180x360x5 visual memory. VM and trajectory are treated as the same latent vector and denoised once, while VAE processes only the visual information.

The visual memory encoding can be decoded to show how visual memory is expected to march forward by the model. Including the visual memory encoding in the diffusion process also provides a foundation for the model to base the predicted trajectory. In practice, we notice that the trajectory prediction samples are more stable with the visual memory prediction.
In inference time, the model will take in the same condition and generate a batch of 15 different samples, forming a discrete distribution of the expected future ego trajectory modes.

We used Smooth L1 loss for diffusion model training. The VAE is trained on the same dataset as the diffusion model. For VAE, a combination of InfoLoss\cite{zhao_infovae_2018}, Cross-entropy loss, and L1 loss is used. InfoLoss makes sure the encoding is highly correlated to the input, L1 loss is performed on RGBD channels, and Cross-entropy is performed on the one-hot semantic channels.


\subsection{Hybrid Generation}

A conventional DDPM sampling method, while capable of producing high-quality predictions, operates at a pace that is impractical for applications requiring immediate responses, such as navigational aids or interactive systems. This limitation is further pronounced when attempting to generate a distribution of future trajectories, as multiple denoising sequences are necessary to produce a substantial number of samples, exacerbating the time constraints. Conversely, while DDIM offers a considerable acceleration in generating predictions, it does so at the expense of sample quality—a compromise that is untenable for applications where the fidelity of predicted trajectories directly impacts functionality and safety.

To address these challenges, we introduce a hybrid generation scheme that synergizes the strengths of both methods. Hybrid generation operates by initiating the generation process with a DDIM-like approach to quickly approximate the trajectory distribution, followed by a refinement phase using the DDPM framework. Essentially retaining the multimodal gradient landscape at the end of the diffusion process, ensuring that the final output maintains the intricate details and nuanced variations captured by a traditional DDPM without the accompanying latency.


\subsection{Diffusion Model Detail}

\begin{figure*}[tb]
  \centering
  \includegraphics[width=1\linewidth]{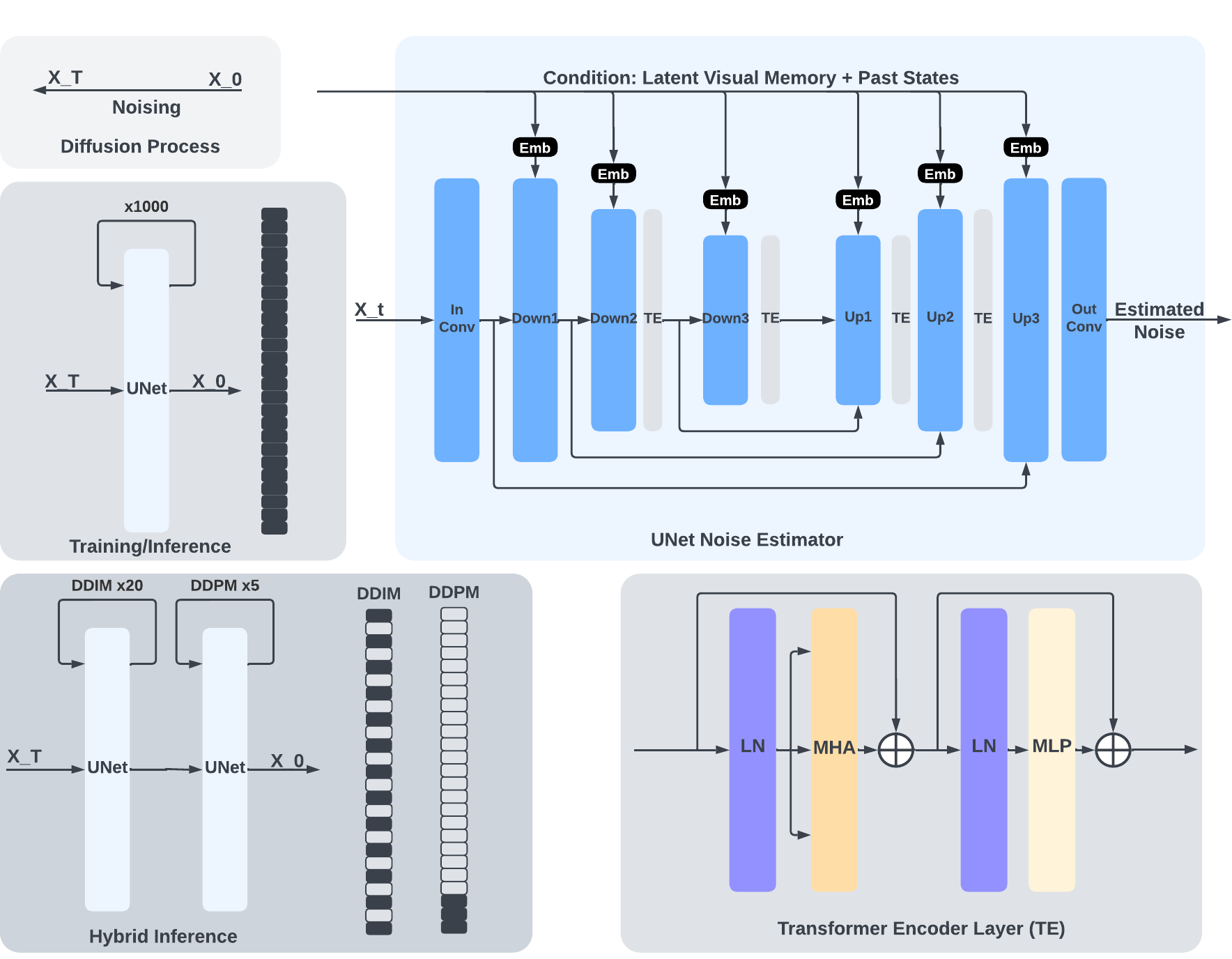}
  \caption{\textbf{Diffusion Model:} Architecture and hybrid generation details. The black and white stripes represent the time-stepping schemes of the inference. In default mode, all steps are diffused sequentially. In our hybrid mode, first perform stripped DDIM steps from 1000 to 0, then DDPM steps from n to 0. Conditions include compressed VM and past states are passed to the embedding layer and concatenated to the block output.}
  \label{fig:model_arch}
\end{figure*}

Fig \ref{fig:model_arch} illustrates the detailed model architecture. We use the transformer encoder blocks which contain multi-head self-attention layers. They are stacked between the Down and Up blocks of the UNet. The conditions of the generation (past trajectory and most recent visual memory) are first turned into embeddings and then passed into the Down and Up blocks.

In training, instead of cutting up the dataset into unique sub-trajectories, we take overlapping sub-trajectories to further augment the dataset. This way, there is a total of more than 220,000 diverse sub-trajectories for the model to train on. 

At inference time, one can either use DDIM, DDPM, or our hybrid inference method. Specifically, the hybrid inference involves first getting a rough convergence with 20 steps of DDIM, the time steps are sub-sampled uniformly from the total steps. Then the last 5 steps are performed with DDPM steps 5 to 0 to bring the sample to full convergence. The variance we used in DDPM is fixed small, or mathematically:

\begin{align}
    x_{t-1} = \frac{1}{\sqrt{\alpha_t}}(x_t - \frac{1-\alpha_t}{\sqrt{1-\Bar{\alpha}_t}}z_{\theta}(x_t,t))+\sigma_tz
\end{align}
where:
\begin{align}
    \sigma_t = \frac{1-\Bar{\alpha}_{t-1}}{1-\Bar{\alpha}_t}\beta_t
\end{align}

\section{Egocentric Navigation Dataset}
\label{sec:dataset}

\begin{figure*}[tb]
  \centering
  \includegraphics[width=\linewidth]{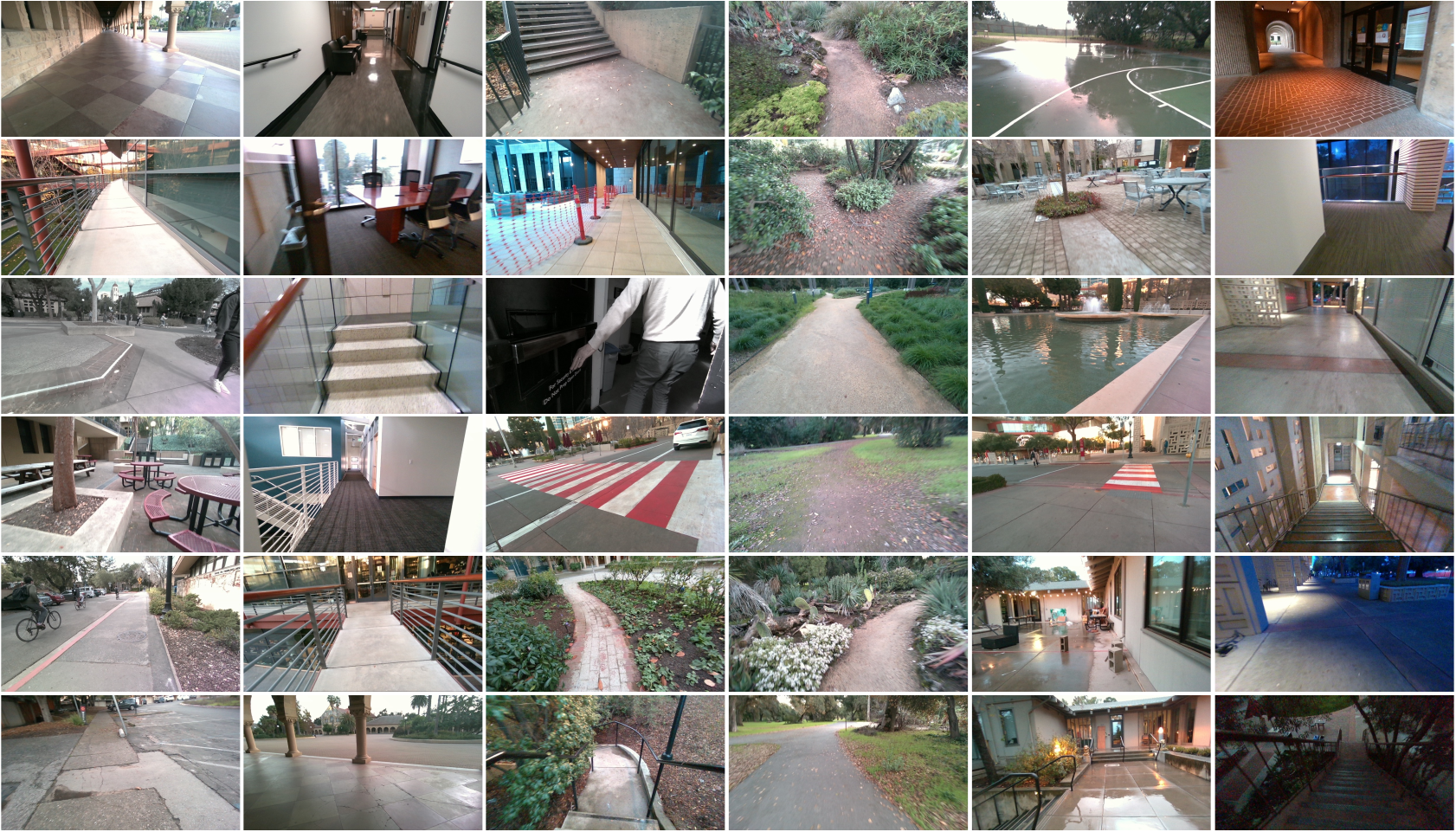}
  \caption{\textbf{Dataset:} The dataset contains a healthy mix of weather, road, lighting, and traffic conditions}
  \label{fig:traj_features}
\end{figure*}

Our Egocentric Navigation Dataset is collected within a university campus and its vicinity on human participants (under the approval of IRB-60675). The dataset comprises 34 carefully selected collections, each lasting approximately 7 minutes and spanning over 600 meters, designed to capture a wide range of interactions with the environment. These interactions are critical for testing and enhancing trajectory prediction models, particularly in densely populated or complex areas. The dataset is characterized by its diversity, encompassing various weather conditions (rain, sunny, overcast), surface textures (glass, solid, glossy, reflective, water), and environmental features (stairs, ramps, flat grounds, hills, off-road paths), alongside dynamic obstacles, including humans. Such variety ensures the dataset's applicability across a wide spectrum of egocentric navigation and prediction tasks. It is worth noticing that we did not deliberately avoid any pedestrians, reflecting real-world campus walking scenarios.

Recorded at a 20Hz sampling rate, the dataset includes comprehensive state and visual information to capture the nuances in human behavior: 6 degrees of freedom (dof) torso pose in a global frame, leg joint angles (hips and knees of both legs), torso linear and angular velocity, and gait frequency. Visual data comprise aligned color and depth images, semantic segmentation masks, and visual memory frames generated to aid in trajectory prediction. This extensive collection amounts to 198 minutes of data (over 400GB). In practice, we find it is possible to train a very high-quality model even with a smaller dataset size. Therefore we curated a high-quality pilot dataset with roughly 15\% of the full data. This allows us to quickly iterate with faster training and an acceptable amount of performance degradation. The performance has been qualitatively compared in Sec \ref{sec:evaluation}.

Previous datasets have often suffered from sparse content, low logging frequencies, and lack of access to raw data or pre-trained decoder weights, hindering the development of sophisticated models and methods. Acknowledging the limitations of existing datasets in this domain, such as the TISS dataset\cite{qiu_egocentric_2022}, our Egocentric Navigation Dataset addresses critical gaps by providing dense, high-frequency logs with rich visual and state information. Recently, Meta released the Aria Everyday Activities Dataset (AEA)\cite{lv_aria_2024}, which also adopts an ego-centric view. Our dataset offers distinct advantages for outdoor navigation scenarios by providing dense depth information from sensors. In contrast, AEA focuses on indoor scenes and daily manipulation activities, offering SLAM key points rather than dense depth maps. While both datasets contribute to egocentric research, ours is particularly well-suited for tasks requiring detailed depth perception and outdoor navigation, allowing researchers to choose the most appropriate resource for their specific research needs.

To foster innovation and accessibility in egocentric trajectory prediction research, we are committed to open-sourcing our dataset following the de-identification of all faces within the data. We believe this approach will democratize access to high-quality data, enabling researchers to explore new methodologies and applications in the field.

    \begin{figure*}[tb]
      \centering
      \includegraphics[width=\linewidth]{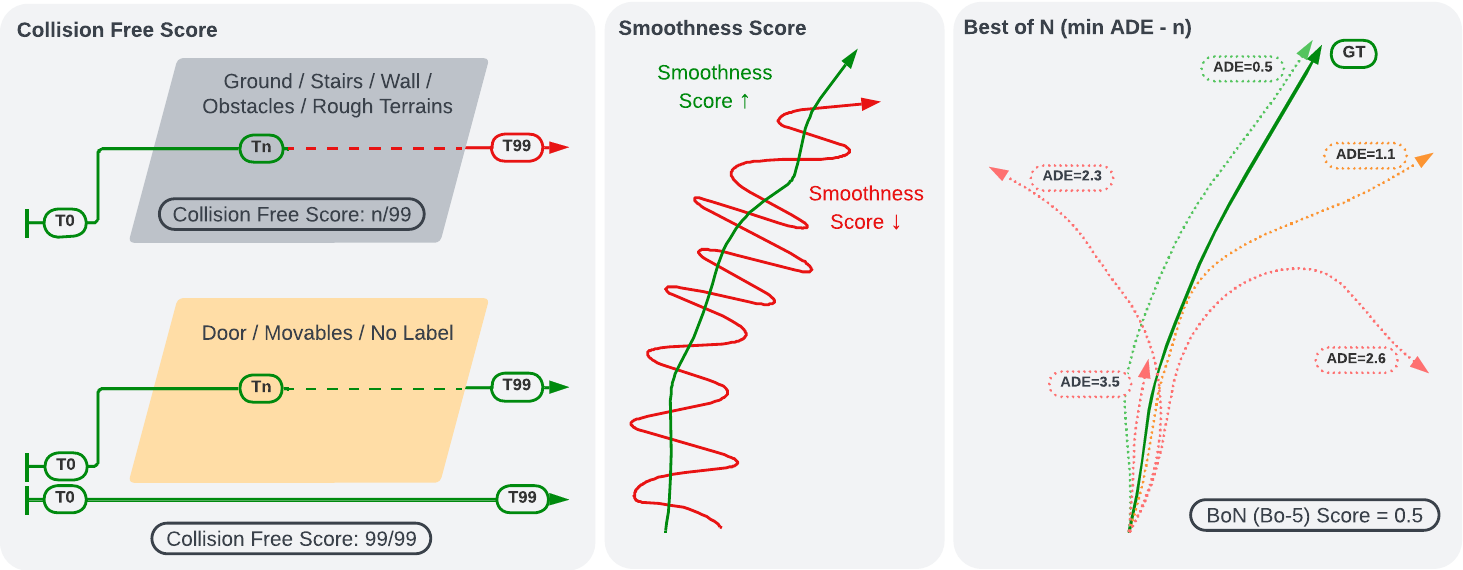}
      \caption{\textbf{Overview of metrics:} The collision-free score (CFS) only considers points with selected semantics. All subsequent trajectories will be marked as collided. CFS and Smoothness are both higher the better, as opposed to Best of N.}
      \label{fig:overview_metrics}
    \end{figure*}
    
\subsection{Dataset Detail}
All the data are collected at a rate of 20Hz, both state and visual inputs. State information includes 6 dof torso localization in the global frame, leg joint angles ({left, right} calf and thigh), torso velocity, torso angular velocity, and average gait frequency in 2 seconds moving window. Visual information includes RGB images, an aligned depth frame from D455 at a resolution of 848x480, semantic segmentation masks by DINOv2 + Mask2Former segmentation head, as well as panoramas with all the aforementioned channels projected to 360 view. This brings the total dataset to 198 minutes, and over 400GB. 

We optimize storage by retaining raw depth frames only when there's a significant change in camera position—specifically, an angular movement exceeding 15 degrees or a translation beyond 1 meter. This strategy reduces the point cloud processing rate to approximately 1 keyframe per second at normal walking speeds, thus conserving computational resources for semantic segmentation. Visual memories are generated in a separate process at 20Hz, center-aligned with the current camera pose, to support timely model inference.

We found that the raw data from stereo cameras requires prepossessing to enhance quality, as it tends to connect disjoint edges in the scene. We apply a tuned Canny edge filter to mitigate artifacts along object edges, removing up to 10 pixels around the disjoint edges, this prepares the depth frames for more accurate panorama construction.

We define eight semantic classes for the segmentation: No label, normal ground, stair, door, wall, obstacle, movable, and rough ground. The collision evaluation framework specifically considers ground, stairs, walls, obstacles, and rough terrains to accurately differentiate between navigable and non-navigable spaces.

\section{Evaluation}
\label{sec:evaluation}
    \subsection{Hardware Setup}
    We utilized a similar hardware setup as this previous work \cite{wang_trajectory_2023} and made some upgrades. To accommodate data collection in both indoor and outdoor environments, we opted for a setup that does not rely on GPS due to its indoor limitations and avoids pure IMU localization to minimize drift over extended periods. For precise localization, we employ an Intel Realsense T265 for SLAM-based tracking. We upgraded the stereo camera to an Intel Realsense D455, benefiting from its longer stereo baseline to enhance depth perception accuracy and extended the observable range. Considering the variability in lighting conditions, especially outdoors, we chose not to use LiDAR, despite its accuracy in optimal lighting conditions. The entire software stack, including real-time panorama generation and data storage, and physically powering the sensor is handled by an Apple Silicon MacBook Pro housed in a backpack. This setup maximizes mobility and flexibility during the data collection.
    \subsection{Evaluation Metrics}

    To comprehensively evaluate the predicted trajectories, we introduce three key metrics: collision-free score, smoothness, and best of N (including best of 1 as a subset for comparison with unimodal predictions). These metrics are designed to reflect the multifaceted nature of trajectory prediction, capturing aspects of physical feasibility, motion quality, and alignment with potential human preferences. An infographic summarizing these metrics is depicted in Fig \ref{fig:overview_metrics}.
        \subsubsection{Collision-free Score}
        This metric assesses the feasibility of the predicted trajectories by ensuring they do not intersect with impermeable objects within the environment. Due to the absence of ground truth environmental meshes, we evaluate collisions against a discrete point cloud. To calculate the collision-free metric, we first re-project the visual memory back into a 3D point cloud. At each predicted time step, we retrieve the closest 20 points in the point cloud using a K-D Tree. We define a collision as the presence of more than 10 scene points within a 16cm radius of the predicted position. If a collision is detected, the current and all subsequent time steps are deemed collided. The collision-free score tallies the number of consecutive prediction steps where the trajectory avoids collision with static obstacles such as the ground, stairs, walls, and rough terrains, excluding doors and movable objects. The metric value is the index of the first collided time step, with a maximum score of n-1 for n prediction steps. A higher score indicates better performance. The radius (16cm) and threshold (10 points) parameters are tunable, and these empirical values were found to be most reasonable for our resolution.

    \begin{table*}[tb]
      \caption{Metric comparison table across ablations. Bold = Best, Italic = 2nd}
      \label{tab:ablations}
      \centering
      \begin{tabular}{@{}l|cccc@{}}
        \toprule
        Ablation & Collision ↑& Smoothness ↑& Best of 1 ↓& Best of 15↓\\
        \midrule
        Ground Truth & 97.6 & - & - & - \\
        Ours (Full dataset)     & \textbf{90.6} & \textbf{4.76}  & \textbf{0.81}  & \textbf{0.41} \\
        w/o visual memory prediction & \textit{89.3} & 3.13 & 0.91 & 0.50 \\
        Ours      & 89.2  & 2.04 & \textit{0.87}  & \textit{0.47}\\
        w/ Markovian past state     & 88.8 & 1.56 & 1.04 & 0.52 \\
        w/ Hybrid generation (DDIM20+DDPM10)     & 88.7 & 2.17 & 0.89 & 0.49\\
        w/o attention     & 86.6 & 2.78 & 1.00 & 0.49 \\
        w/ DDIM generation (n=30) & 85.3 & 0.46 & 0.92 & 0.52\\
        w/o semantic (RGBD only)    & 84.1 & \textit{4.17} & 0.91 & 0.53\\
        w/o visual input (Traj only)     & 82.5 & 2.04 & 1.19 & 0.48 \\
      \bottomrule
      \end{tabular}
    \end{table*}

        \subsubsection{Smoothness}
        The smoothness metric quantifies the continuity and fluidity of movement in the generated trajectories. Calculated as the reciprocal of the sum of the mean absolute errors (MAE) of speed and acceleration compared to ground truth, this metric relates higher values to smoother trajectories. This adjustment ensures that higher scores correspond to desirable attributes, aligning with the intuitive understanding that smoother trajectories are preferable.

        \begin{align*}
            Smoothness = 1/(\frac{\sum^{n}_{i=1}|V_i - \hat{V}_i|}{n} + \frac{\sum^{n}_{i=1}|a_i - \hat{a}_i|}{n})    
        \end{align*}

        \subsubsection{Best of N \& Best of 1}
        Recognizing the inherent multiplicity of valid trajectories for any given scenario, the Best of N metric evaluates the model's ability to encapsulate the ground truth trajectory within its distribution of predictions. This metric is also known as the minADE-K metric widely used in trajectory prediction tasks especially in the autonomous driving field, focusing on the minimum average displacement error across a set of K predictions. The Best of 1 metric, equivalently the ADE metric, complements this by assessing the accuracy of a single prediction, facilitating comparison with models that do not generate a multimodal distribution. It also shows how well the estimated distribution aligns with the ground truth under the assumption that the ground truth mode should have the highest likelihood in the given scenario. Together, these metrics ensure a balanced evaluation of the model's predictive capacity and its versatility in capturing the range of plausible ego-motions.

    \begin{figure}[b]
      \centering
      \includegraphics[width=1\linewidth]{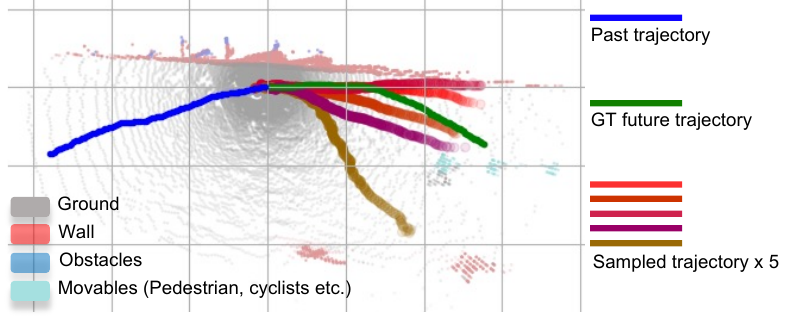}
      \caption{\textbf{BEV:} Typical input and output are visualized top-down. Five trajectories are predicted, each with a different shade of red. Visual memory is projected to a point cloud and color-coded same as semantic.}
      \label{fig:output_topdown}
    \end{figure}
    
    \subsection{Prediction Evaluation}
    Our evaluation focuses on assessing the model's capacity for future trajectory prediction, and its ability to generate expected future visual memory encodings, as detailed in Sec \ref{sec:method}. Operating over a 100-step (5-second) prediction horizon—aligned with the typical visibility range (8-meter radius) and walking speeds—our method ensures the generation of predictions that are both relevant and plausible within the observable environment. 
    An example of output is visualized in Fig \ref{fig:output_topdown}. The figure shows everything in a top-down view, specifically the visual memory given is turned into a point cloud with different colors representing different semantic classes. We can see a wall on the top and two pillars on the bottom, along with some people. The 5 predictions from the model are labeled in red, and the actual ground truth is green. Upon visual inspection, the outputs are diverse and all feasible given the scenario.
    
\begin{figure*}[tb]
      \centering
      \includegraphics[width=\linewidth]{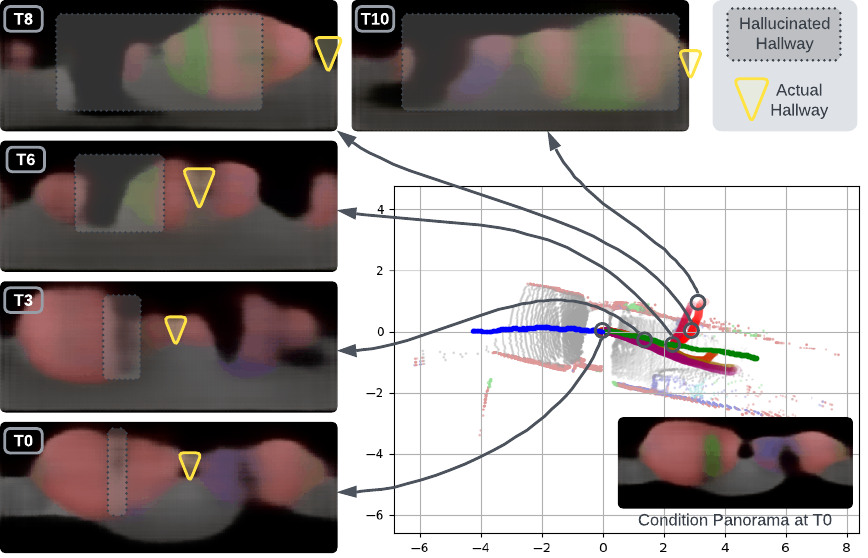}
      \caption{\textbf{Analysis of fail prediction:} A failed prediction (bright red) is picked to analyze the failure mode. Predicted visual memory frames are sampled along the trajectory to show the model's expectation. Visual memory is always center aligned to the forward direction, turning can be observed by following the centerline. BEV plot: The blue trajectory is the past, and the green is the ground truth. Point clouds follow semantic color codes. }
      \label{fig:output_futurama}
    \end{figure*}

    The inclusion of future visual memory alongside trajectory predictions enhances our ability to visually validate the model's anticipations of scene evolution, providing a robust cross reference for trajectory prediction and insight into the model's typical failure modes.
    A pertinent example of this evaluation is depicted in Fig \ref{fig:output_futurama}, where visual memories associated with the red trajectory prediction reveal a challenge commonly encountered by generative models: the occasional "hallucination" of non-existent features in scenarios of occluded vision or gaps in the observable scene. For instance, as a person navigates a hallway, gaps in conditioning visual memories might occur due to changes in view angle or partial occlusions. At a specific timestep (T3), the model inaccurately anticipates that approaching a gap would reveal a larger, unexplored space (as in T6), leading it to "hallucinate" an extension of the hallway based on patterns observed in the training data. This results in the model predicting a trajectory that erroneously veers into this fictitious extension (T8), as evidenced by the gradual shift of the actual hallway from the center of the predicted path.

    \subsection{Ablations}
    To test the effectiveness of individual modules proposed in our method, we conducted the following ablation studies. The individual ablations and corresponding metrics are summarized in Table \ref{tab:ablations}. As mentioned in Sec \ref{sec:dataset}, we have a full dataset for the best possible performance and a curated pilot dataset for quick iteration. All of the ablation training, unless specifically mentioned, are trained on the pilot dataset. The entries in the table have been sorted by the collision-free score in the second column. Notice that the ground truth trajectory did not receive a perfect score (99) in collision avoidance, this is due to the inaccuracies in the SLAM pose estimation and depth artifacts from the stereo camera. In rare occasions, ground truth trajectory collides with a part of the point cloud that appears a considerable distance away from the ground truth surface of the scene.
    

        \textbf{Semantic Segmentation}
        In this task, VAE and diffusion model are trained on the same dataset except without the semantic channel. We see a significant drop in collision avoidance (-5.1). Without semantics, the model will not be able to differentiate between penetrating the door vs the wall. This conflicting behavior in training data makes the model tends to ignore the geometric constraints provided in the visual memory.

        \textbf{Visual Input}
        As expected, the complete removal of the visual input results in the worst-performing model. This emphasizes again the effectiveness of contextual structure provided by the visual memory. Upon inspecting the results, many of the generated samples are rote memorization of the training set and the distribution is largely incorrect (which corresponds to the worst best-of-1 score as explained in Evaluation Metrics of Sec \ref{sec:evaluation}). 
        
        \textbf{Scalability}
        To quantify the difference between training on a large dataset and the pilot dataset, we trained two versions of the baseline one on each dataset. The full dataset model is better on all of the evaluation metrics as expected. However the gap is quite small on the collision metric (+0.6), we believe the collision avoidance performance is more sensitive to the accuracy of the point cloud, than the amount of the training data. If the point cloud quality is low, the model would think that the ground truth demonstration has penetrated some points despite the points are from artifacts and mis-estimations. On the other hand, the added samples have greatly improved the variance of the generation and hence better mode coverage (Best of n -0.06) and best sample quality (Smoothness +2.72)
        
        \textbf{Transformer UNet vs UNet}
        Adding in the multi-head self-attention layers (MHSA) between down and up blocks in UNet significantly blows up the size of the model, fortunately, the inference and training speeds have remained largely unchanged. With attention layers, the most significant boost in performance is in collision avoidance (+4.0). This corroborates our expectation that the MHSA layers help to relate encoded visual memory with trajectory samples.

    \begin{table*}[tb]
      \caption{Metric comparison table across baselines. Bold = Best, Italic = 2nd}
      \label{tab:baselines}
      \centering
      \begin{tabular}{@{}l|cccc@{}}
        \toprule
        Model &  Collision ↑&  Smoothness ↑ &  Best of 1 ↓&  Best of 15 ↓\\
        \midrule
        CXA Transformer & 80.3 & \textit{3.70} & 1.25 & N/A \\
        LSTM-VAE        & \textit{84.5} & \textbf{9.09} & \textit{1.01} & N/A \\
        Ours      & \textbf{89.2} & 2.04 & \textbf{0.87} & \textbf{0.472}\\
      \bottomrule
      \end{tabular}
    \end{table*}

        \textbf{Hybrid Generation \& Computation Efficiency}
        As proposed in Sec \ref{sec:method}, hybrid generation can greatly speed up generation at the expense of a slight decrease in quality. We were able to obtain 50 trajectory predictions per second on a single Nvidia A5000 GPU. In practice, this is enough to give a sizeable distribution covering the majority of the possible cases in the scene. The GPU utilization rate is usually above 95\% during inference. In this experiment, our hybrid generation uses 20 steps of DDIM, followed by 10 steps of DDPM to refine the generation. Compared to a full 1000-step DDPM baseline, this method is 33 times faster, but the quantitative metrics only decreased slightly (Collision -0.5, Best of n +0.02). To show the effectiveness, we also include the same 30-step generation result except for all steps from DDIM. Both collision (-3.9) and smoothness scores (-1.58) were both significantly negatively impacted. This shows that while DDIM is converging in the right direction, it does not converge to as strong a solution as DDPM.

        \textbf{Past Condition Length}
        To achieve the best possible performance, we are using a non-Markovian assumption when performing the prediction, meaning a long history of past trajectories is provided. We are curious as to how well the Markovian assumption holds and are interested in potentially reducing the number of inputs to the model for more efficient use of computation. Therefore we trained a variant that only takes in 3 steps of past trajectory steps, essentially only providing information up to the order of acceleration. The result: Collision only slightly decrease (-0.4), since collision performance is mostly based on visual information; Smoothness is notably worse than before (-0.48), as the model receives less information about the momentum; Mode coverage is impacted slightly (Best of N +0.05). This shows that the Markovian assumption largely holds for most of the scenarios. And we can potentially speed up more using this model in deployment.

    \subsection{Comparison with Baselines}
    The proposed method is compared to the current state-of-the-art methods: LSTM and CXA Transformer. For consistency, all three methods are based on the same pre-trained VAE. And they are all trained and evaluated on the pilot dataset. Results are compiled in Table \ref{tab:baselines}. We would like to highlight some key differences between our model and the baselines: 1) Our model does not rely on auto-regressive generation and, therefore less affected by compounding error problems. Thus also requires less pass through the generation model, meaning faster speed. 2) Our model is inherently based on a random process, therefore it learns an implicit distribution of preferences instead of simply an average.
    
    
        \textbf{VAE-LSTM}
        One of the most common paradigms to the trajectory prediction task in the existing literature \cite{wang_trajectory_2023, tang_multiple_2019, schreiber_long-term_2019, su_predicting_2017, varadarajan_multipath_2022} is to use an encoder head to understand the visual input, and then pass the state information along with encoded visual cues into a recurrent model to auto-regressively predict the future states. The biggest drawback of this approach is that the prediction is deterministic, as the variance only exists in the encoding process of the visual input, not in the trajectory latent space. And since there is only one deterministic prediction in each scenario, there is zero guarantee of the temporal consistency of the prediction between consecutive time steps. Despite the drawbacks, the auto-regressive model excels at generating smooth motion as every step is based on the previous predictions, which is also confirmed by the smoothness score (+7.05). The LSTM baseline performs surprisingly well at avoiding obstacles at a tiny size of fewer than 1 million parameters.
        
        \textbf{CXA-Transformer}
        The more recent approach is to use the transformer instead of the recurrent modules\cite{qiu_egocentric_2022, mercat_multi-head_2019}. The transformer models, at its core, rely on attention mechanisms to process the sequence so it suffers less from forgetting and improves computational efficiency for scaling. One of the most recent works\cite{qiu_egocentric_2022} aiming to tackle egocentric trajectory prediction proposes a Cascaded Cross-Attention(CXA) Transformer. It uses a transformer encoder for each modality of input including surrounding people's pose, semantic segmentation, and past trajectories, and a cross-attention module to fuse in multi-modal information. Similar to other transformer models, it uses a transformer decoder to produce a prediction of the most likely future path auto-regressively. 
        CXA transformer is implemented strictly as outlined in the paper. To optimize its performance on our dataset, we started from the provided hyper parameters and did some tuning. Notably, the collision score of the CXA transformer is significantly lower (-8.9), potentially because it uses stereo camera frames directly with a limited field of view. Unfortunately, the CXA transformer also only returns a single modal prediction. Despite that, the Best of 1 score is still worse than our baseline (+0.38). As an auto-regressive model, the smoothness is still its strength (+1.66).

        This showcases the unique advantage of our method using the diffusion model for multi-modal prediction, and Visual Memory to maximize the situational awareness of the model.

\section{Conclusion}
\label{sec:conclusion}
This paper presented a novel pipeline to predict future trajectories using a diffusion model conditioned on the past trajectory and scene semantics. In the process, we proposed an efficient representation called Visual Memory, a new sampling scheme to speed up sample generation for real-time prediction, and a diverse egocentric human walking dataset to enhance future research in the field. We showed that our method is highly effective through comprehensive ablations and comparisons with current state-of-the-art baselines. 

Future work in this area includes employing monocular depth estimation models to relax the need for the stereo depth camera. Distilling the latent diffusion model to consistency model \cite{song_consistency_2023} for real-time generation is also desirable especially when deploying the model to mobile platforms.

\bibliographystyle{IEEEtran}
\bibliography{KenBib}

\end{document}